\documentclass[letterpaper, 10 pt, conference]{ieeeconf}  
\usepackage{graphicx} 
\usepackage{amssymb}
\usepackage{amsmath}
\usepackage[dvipsnames]{xcolor}

\IEEEoverridecommandlockouts                              

\title{\LARGE \bf
3D Object Visibility Prediction in Autonomous Driving
}

\author{Chuanyu Luo$^{1}$, Nuo Cheng$^{1}$, Ren Zhong$^{2}$, Haipeng Jiang$^{2}$, Wenyu Chen$^{3}$, Aoli Wang$^{3}$, Pu Li$^{3}$ 
\thanks{*This work is supported by LiangDao GmbH}
\thanks{$^{1}$Chuanyu Luo, Nuo Cheng are with the LiangDao GmbH, Berlin, 12099, Germany and also with the Ilmenau University of Technology, Ilmenau, 98693, Germany {\tt\small chuanyu.luo@tu-ilmenau.de; nuo.cheng@tu-ilmenau.de}.}%
\thanks{$^{2}$Ren Zhong, Haipeng Jiang are with the Great Wall Motor Co., Ltd. {\tt\small reazn@qq.com; 18531225980@163.com}.}%
\thanks{$^{3}$Wenyu Chen, Aoli Wang, Prof. Pu Li are with the Ilmenau University of Technology, Ilmenau, 98693, Germany {\tt\small  wenyu.chen@tu-ilmenau.de; aoli.wang@tu-ilmenau.de; pu.li@tu-ilmenau.de}.}%
}

\begin{document}

\maketitle
\thispagestyle{empty}
\pagestyle{empty}

\begin{abstract}

With the rapid advancement of hardware and software technologies, research in autonomous driving has seen significant growth. The prevailing framework for multi-sensor autonomous driving encompasses sensor installation, perception, path planning, decision-making, and motion control. At the perception phase, a common approach involves utilizing neural networks to infer 3D bounding box (Bbox) attributes from raw sensor data, including classification, size, and orientation.

In this paper, we present a novel attribute and its corresponding algorithm: 3D object visibility. By incorporating multi-task learning, the introduction of this attribute, visibility, negligibly affects the model's effectiveness and efficiency. Our proposal of this attribute and its computational strategy aims to expand the capabilities for downstream tasks, thereby enhancing the safety and reliability of real-time autonomous driving in real-world scenarios.

\end{abstract}

\section{INTRODUCTION}

In the domain of autonomous driving perception, the prediction of 3D bounding boxes (Bbox)~\cite{lang2019pointpillars}\cite{yan2018second}\cite{liang2022bevfusion} has emerged as the predominant approach over 2D Bbox prediction~\cite{redmon2016you}\cite{duan2019centernet}, which relies solely on images. The key advantage of 3D Bbox prediction lies in its ability to provide critical additional information, such as the distance and orientation of objects. This not only offers redundancy in information but also significantly enhances safety for downstream applications, including path planning and decision-making processes.

In 3D autonomous driving datasets, such as KITTI~\cite{geiger2012we} and nuScenes~\cite{caesar2020nuscenes}, various state-of-the-art neural network perception algorithms~\cite{lang2019pointpillars}\cite{yan2018second}\cite{liang2022bevfusion}  have been developed to predict 3D bounding box (Bbox) attributes, including classification, size, and orientation, directly from raw sensor data. In this paper, we define the process of analyzing raw sensor data to determine these attributes as the \textbf{prediction} stage.

After generating Bbox from prediction stage, information extraction is performed at the Bbox level. This, e.g., includes predicting Bbox velocity and unique IDs using multiple frames~\cite{ma2023detzero} and tracking algorithms~\cite{yin2021center}\cite{wang2021immortal}. In this paper, we refer to the analysis and processing of the Bbox after its initial generation as the \textbf{post-prediction} stage. Therefore, the perception phase encompasses both the prediction and post-prediction stages.

Clearly, if the perception phase can provide high-quality additional information for downstream tasks, such as path plan and vehicle control, without compromising its accuracy and efficiency, it can enhance the safety of autonomous driving.

In this paper, we introduce a \textbf{visibility} attribute and its associated algorithm, which rely solely on 3D Bbox. By transitioning the visibility prediction from the post-prediction stage to within the prediction stage itself, we aim to enhance the accuracy of visibility information without affecting the original prediction-stage's effectiveness and speed, and provide high accurate visibility information for downstream tasks.

In previous studies, KITTI~\cite{geiger2012we} and nuScenes~\cite{caesar2020nuscenes} provide ground truth information for 2D Bbox visibility/occlusion based on images, using the Intersection over Union (IOU) metric. However, when it comes to the prevalent methodologies for 3D prediction, reliance on 2D visibility metrics presents several limitations. These include a dependency on the specific positions at which sensors or cameras are installed, as well as an inability to ascertain visibility values for bounding boxes that fall outside the camera's field of view.

Moreover, within the KITTI and nuScenes datasets, visibility/occlusion plays a role in assessing algorithmic precision, typically by categorizing objects into easy, moderate, and hard levels. However, this feature has been underutilized by both the academic and industrial sectors as a source of supplementary predictive data. Such information could serve as a valuable layer of redundant safety for autonomous vehicles, ensuring a higher level of reliability in their operational environments.

Therefore, our contributions in this paper can be summarized as follows:

1. \textbf{An innovative visibility definition and algorithm}: We introduce a novel definition of visibility and a corresponding algorithm that relies exclusively on 3D bounding boxes (Bboxes). This approach is unique because it is sensor-agnostic, allowing it to be applied across different perception tasks, including those involving solely LiDAR~\cite{lang2019pointpillars}\cite{yan2018second}, purely image-based methods~\cite{li2022bevformer}, and LiDAR-Image fusion~\cite{bai2022transfusion}\cite{liang2022bevfusion}. Our visibility metric is algorithmically determined, which distinguishes it from other Bbox attributes such as classification, size, and orientation that typically require manual annotation. This characteristic enhances its utility in diverse applications without necessitating additional manual input.

2. \textbf{Neural networks and multi-task learning integration}: Through the integration of neural networks and multi-task learning strategies, we empirically show that shifting the visibility calculation to the prediction phase in LiDAR-based perception tasks enhances the accuracy of visibility information for subsequent processes. This shift does not compromise the accuracy or speed of perception, thereby contributing to the safety of autonomous driving systems in real-world scenarios. Our findings suggest that this methodology can significantly improve the efficiency of perception systems by providing crucial visibility information early in the process, potentially enhancing the overall safety and reliability of autonomous vehicles.

\section{RELATED WORK}
In this section, we begin by examining previous models of 3D object detection. Subsequently, we delve into research related to multi-task learning. Finally, we explore the role of visibility in the context of autonomous driving.

\subsection{3D object detection models}

In the field of LiDAR-based 3D object detection, the pioneering approach, PointNet~\cite{qi2017pointnet}, is introduced to directly process the raw point cloud data for tasks such as 3D indoor object classification and segmentation. Given the inherently unordered nature of point cloud data, PointNet proposes a symmetric function, specifically max-pooling, to efficiently aggregate information from the unordered point cloud.

In outdoor, large-scale, LiDAR-based driving scenarios, VoxelNet~\cite{zhou2018voxelnet} introduces a method to segment an irregular point cloud into a uniform grid of dense 3D voxels. It then employs a dense 3D Convolutional Neural Network (3D-CNN) as the foundational architecture for its object detection model. However, point clouds in outdoor environments are typically sparse. Addressing this issue, SECOND~\cite{yan2018second} presents a novel approach by proposing a sparse 3D convolutional backbone designed for real-time 3D object detection. Unlike traditional methods, SECOND's convolutional kernels focus on aggregating features exclusively from voxels that contain points, optimizing processing efficiency and detection performance in sparse point cloud environments.

Inspired by PointNet and SECOND, PointPillars~\cite{lang2019pointpillars} has been developed to segment a point cloud into 2D pillars, conceptualized as vertical columns of points. This approach has markedly enhanced the model's efficiency, while only causing a negligible decrease in performance.

Recent advancements in 3D object detection have harnessed the capabilities of Transformer~\cite{vaswani2017attention}, leading to significant innovations in both image-only and LiDAR-image fusion methodologies. BEVformer~\cite{li2022bevformer}, designed for image-only contexts, employs Transformers to convert multi-camera inputs into bird’s-eye-view (BEV) features, offering a novel perspective on scene interpretation. In the scope of LiDAR-image fusion, Transfusion~\cite{bai2022transfusion} integrates Transformers within its LiDAR and image decoders, enhancing the model's ability to selectively utilize image-derived information. Similarly, BEVfusion~\cite{liang2022bevfusion} leverages Transformer technology to efficiently align features from multiple sensors within a unified BEV representation space, facilitating a more integrated approach to sensor fusion.

\subsection{Multi-task learning in autonomous driving}
Multi-task learning aims to enhance both efficiency and accuracy by leveraging shared information across multiple tasks~\cite{liang2022effective}. MultiNet~\cite{teichmann2018multinet} employs a unified architecture to tackle tasks in image classification, detection, and semantic segmentation. In the field of LiDAR technology, LidarMTL~\cite{feng2021simple} processes raw LiDAR point clouds to generate six perception outputs, facilitating 3D object detection and road understanding. Similarly, LidarMultiNet~\cite{ye2023lidarmultinet} utilizes a single, comprehensive network to concurrently execute tasks in 3D semantic segmentation, 3D object detection, and panoptic segmentation.

Note that the aforementioned studies necessitate multi-task ground truth data derived from human labeling. In contrast, our work distinguishes itself as the ground truth for our auxiliary tasks does not depend on manual annotations but is instead generated algorithmically.

\subsection{Visibility in 3D object detection}

In the task of 3D object detection,~\cite{hu2020you} contends that point cloud data does not constitute genuine 3D data because of occlusion, effectively rendering LiDAR data as 2.5D. By integrating occlusion information, which is calculated via ray casting, into point clouds, there is a notable enhancement in the accuracy of 3D object detection.

The closest paper to our work is~\cite{chu2022visibility}, which investigates Monocular 3D object detection, aiming to predict 3D objects from two-dimensional images. In the field of monocular 3D object detection, it suggests that occlusion phenomena significantly limit performance in practice. It achieves this by defining and predicting visibility information for the eight key-points of the Bbox . 

In this study, we introduce a novel definition of visibility that more closely aligns with human perception. We investigate its effectiveness in a purely 3D environment, utilizing LiDAR technology. Our approach employs multi-task learning to demonstrate that this new visibility metric achieves accuracy comparable to traditional algorithmic calculations, with only a minimal impact on the overall real-time model's performance and processing pipeline.

\section{METHOD}

In this section, we delve into the methodology underpinning our study, detailing it through a structured approach that includes the following key steps:

\begin{enumerate}
  \item \textbf{Exploring the definitions of visibility:} We commence by elucidating  various potential interpretations of visibility, setting the stage for a comprehensive understanding of its conceptual framework.
  \item \textbf{Algorithm development for visibility calculation:} Following the establishment of our visibility definitions, we introduce a meticulously designed algorithm. This algorithm is tailored to compute the visibility as defined, ensuring accuracy and relevance in our approach.
  \item \textbf{Integration of multi-task learning for enhanced precision:} By integrating multi-task learning techniques, we significantly enhance the precision of our visibility calculations. This advancement is achieved with minimal impact on computational speed, striking a balance between accuracy and efficiency.
\end{enumerate}

\subsection{Definition of visibility}

Contrary to widely recognized attributes like Bbox dimensions, orientation, and others, the concept of visibility—or conversely, occlusion—lacks a predefined or universally accepted characterization. Hence, establishing a clear definition is essential. In this study, we consider three potential definitions for visibility:

\begin{enumerate}
    \item For the KITTI and nuScenes datasets, visibility determination involves projecting 3D Bboxes onto the corresponding camera's view. This process is followed by calculating the visibility metric through 2D mIoU on the related image.
    \item Within the domain of LiDAR, visibility assessment accounts for the interaction between LiDAR beams and the positioning or occlusion of objects. A method to quantify visibility is by measuring the ratio of points within the bounding box that comprehensively fill the Bbox. This approach leads to the observation that objects obscured by obstructions in close proximity exhibit reduced visibility. Conversely, objects at a distance may also display diminished visibility due to the decreasing density of the point cloud, even if they are not occluded by other objects.
    \item In this paper, we suggest projecting 3D Bbox onto a (unit) spherical plane centered at the origin. This method allows for the assessment of Bbox occlusion and mIOU on the spherical plane, akin to evaluations done on a 2D image. It is important to note that, in contrast to conventional images or range images, the projected Bbox on the spherical plane does not retain a rectangular shape, thus introducing a unique perspective for visibility analysis.
\end{enumerate}

\begin{figure}[thpb]
      \centering
      \includegraphics[scale=0.26]{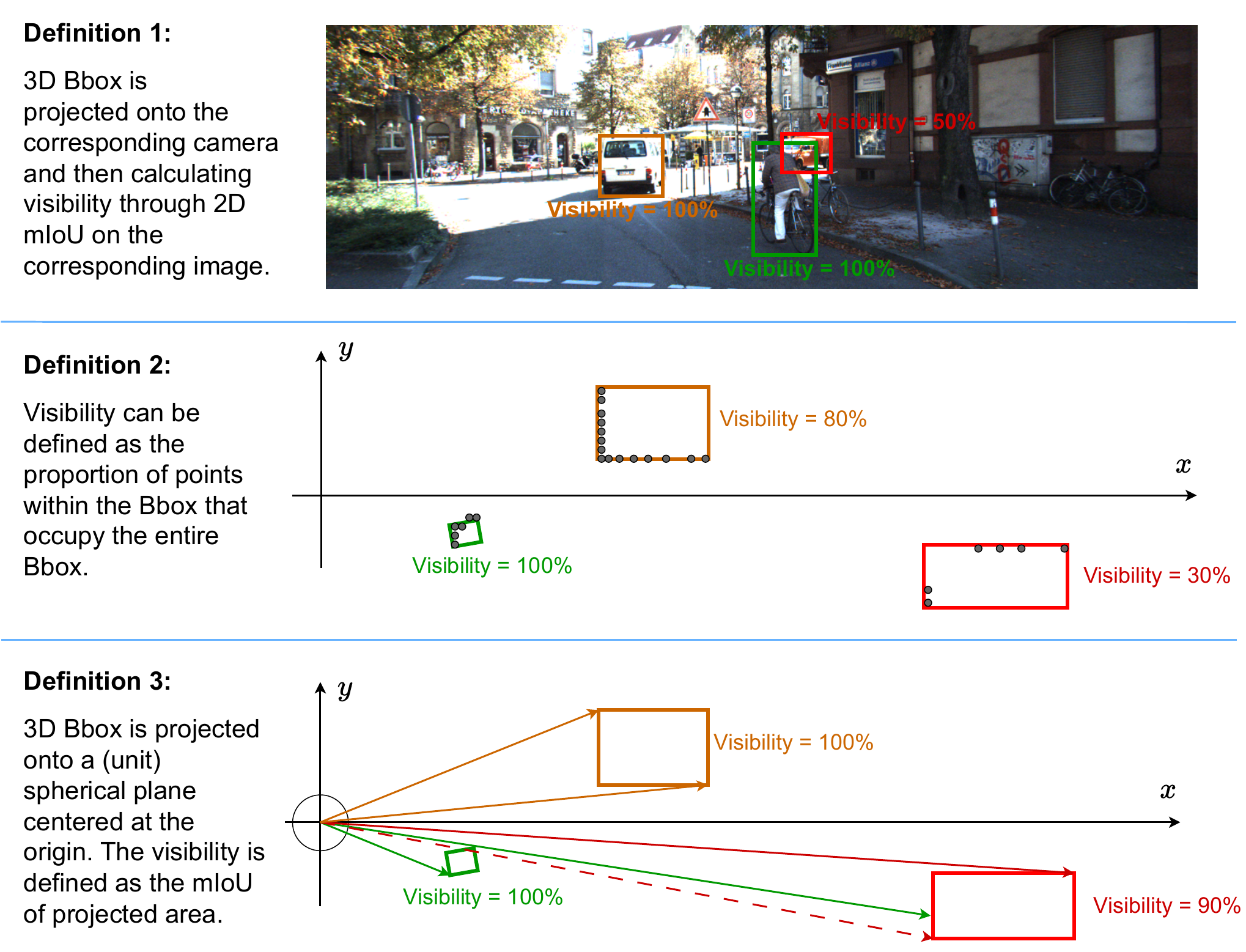}
      \caption{This figure presents three distinct definitions of visibility. For illustrative purposes, Definitions 2 and 3 are depicted from a top-down perspective. However, it's important to note that within our study, both of these definitions are applied and analyzed in a three-dimensional context.}
      \label{visibility_def}
   \end{figure}

These definitions are further illustrated in Figure~\ref{visibility_def}. In contrast to the initial two definitions of visibility, our proposed third definition offers the following advantages:

\begin{itemize}
    \item Calculating the visibility requires only the information from 3D Bboxes. Definition 1 necessitates transformations involving the camera and its corresponding coordinate system, whereas Definition 2 is applicable exclusively to domains utilizing LiDAR technology. Considering that contemporary research and applications in this algorithmic field span pure LiDAR~\cite{lang2019pointpillars}\cite{yan2018second}, solely image-based~\cite{li2022bevformer}, and hybrid LiDAR-Image fusion~\cite{bai2022transfusion}\cite{liang2022bevfusion} methodologies, our introduced Definition 3 offers a more adaptable and comprehensive framework.
    \item Definition 3 entails projecting the bounding box (Bbox) onto a spherical plane, as opposed to a traditional rectangular image format. This approach to design, utilizing a spherical plane, aligns more closely with the mechanisms of human perception.
\end{itemize}

Definition 3 also presents limitations, primarily due to its focus on the occlusion relationships solely among 3D Bboxes (positive objects), overlooking how negative objects (background elements) like buildings or trees may obscure these bounding boxes. Consequently, occlusions by such negative objects are disregarded under Definition 3. Nevertheless, our experimental findings and practical experiences suggest that such occurrences are infrequent, given that Bboxes situated on roads are rarely obscured by background objects.

\subsection{Algorithm to calculate the visibility}
\label{subsec:algo}

In this subsection, we present the algorithm for calculating the defined visibility metric. Initially, we introduce the concept of the solid angle, which quantifies the extent of the field of view from specific points. Within the scope of our study, the solid angle is defined as the field of view from any 3D Bbox towards the origin of the ego-vehicle.

In this context, any 3D Bbox is projected onto the unit sphere centered at the ego-vehicle's origin. The solid angle, denoted as $\Omega$, is mathematically defined as

\begin{equation}
    \centering
\Omega = \frac{A}{r^2}.
\end{equation}

Here $A$ is the area of Bbox projection surface $S$, and $r$ is the radius of projected sphere.

In spherical coordinates, the area $A$ of a projection $S$ on unit sphere can be calculated as
\begin{equation}
    \centering
    A = \iint_S sin\theta \,d\theta\,d\phi
\end{equation}
Here $\theta$ is the latitude and $\phi$ is the longitude.

As a Bbox consists of 6 faces, and each face is projected, the projection of Bbox $A$ is the union of its 6 faces projections, i.e. $A = \bigcup_{i=1}^{6} A_{i}$.

\begin{figure}[thpb]
      \centering
      \includegraphics[scale=0.6]{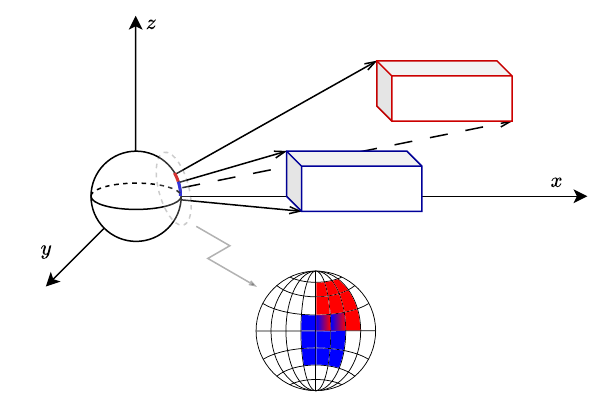}
      \caption{The algorithm for calculating visibility involves projecting the red and blue Bboxes onto a unit sphere centered at the origin. The visibility of the blue Bbox is 100\%, as no other Bbox is closer to the origin. The unoccluded area of the red Bbox is determined by subtracting the area of overlap with the blue Bbox, which is characterized by a gradient color, from its total projected area. Consequently, the red Bbox is partially occluded. Its visibility is calculated as the ratio of the area exclusively occupied by the red projection to the entire projection area.}
      \label{visibility_algo}
   \end{figure}

The visibility $V$ of $i$-th 3D Bbox can be defined as 

\begin{equation}
    \centering
V_i = \frac{\Omega_i - \Omega_i \bigcap (\bigcup_{j=1}^{C_i} \Omega_{j})}{\Omega_i}.
\end{equation}
Here $C_i$ indicates the Bboxes set which are closer to the ego vehicle than the $i$-th Bbox . If no Bbox is closer, the visibility of $i$-th Bbox is 100\%. The projection and occlusion method is further illustrated in Figure~\ref{visibility_algo}. Note that the complexity of the algorithm  is $O(N^2)$, here $N$ is the number of Bboxes.

\subsection{Visibility prediction by multi-task learning}
\label{subsec:visibility_p}

\begin{figure}[thpb]
      \centering
      \includegraphics[scale=0.75]{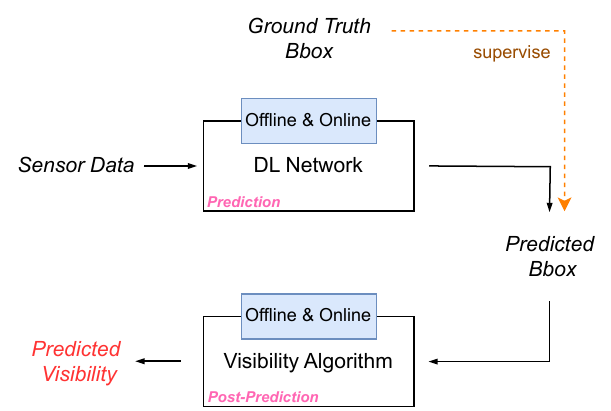}
      \caption{The traditional workflow to calculate the visibility. In the prediction stage, a deep learning model, such PointPillars~\cite{lang2019pointpillars} or SECOND~\cite{yan2018second}, can be used to predict the Bbox. Then the predicted Bbox can be used by proposed algorithm to calculate its visibility in the post-prediction stage.}
      \label{workflow_traditional}
   \end{figure}

In the context of autonomous driving, one traditional approach for assessing the visibility of Bbox involves two primary steps. Initially, a deep learning neural network predicts the Bbox from sensor data. Subsequently, the algorithm introduced in the preceding subsection is employed to determine the visibility of this Bbox. This established workflow is illustrated in Figure~\ref{workflow_traditional}.

In the traditional workflow, visibility prediction or calculation significantly impacts perception phase latency, extending the time required for perception compared to the original no-visibility-workflow. It is important to note that the complexity of the algorithm is $O(N^2)$, where $N$ represents the number of Bboxes. By optimizing with the scan line algorithm~\cite{blinn1978scan},  the complexity can be reduced to $O(NlogN)$. Nonetheless, this level of complexity remains a concern in real-time autonomous driving scenarios. For safe driving practices, the optimal complexity is aimed to be $O(1)$.

\begin{figure}[thpb]
      \centering
      \includegraphics[scale=0.7]{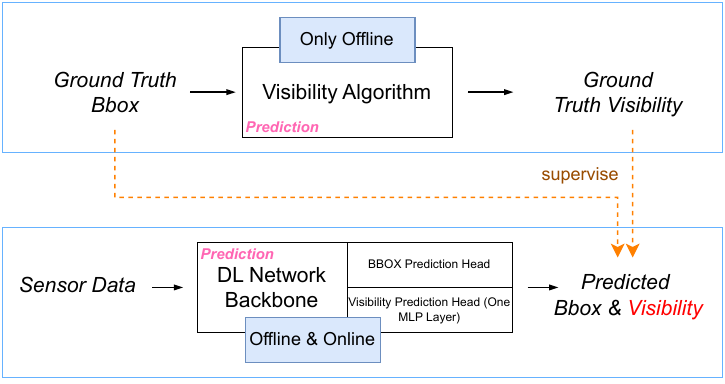}
      \caption{The proposed workflow to calculate the visibility. The visibility is calculated by algorithm offline and used for training in multi-task leaning. In the real-time inference, the visibility is predicted by extending an extra simple MLP layer.}
      \label{workflow_new}
   \end{figure}

In this study, we enhance the predictive capabilities of our model by incorporating multi-task learning within neural networks. This approach shifts the emphasis from post-prediction to the prediction stage itself, significantly improving the model's efficiency with high accuracy. The details of this proposed workflow are illustrated in Figure~\ref{workflow_new}.

Compared to the traditional method, our proposed workflow introduces only a single additional Multilayer Perceptron (MLP) layer to predict the visibility, requiring minimal additional time. Hence, this approach achieves the desired complexity of $O(1)$. 

In our LiDAR experiments shown in the next secion, we demonstrate that this method yields high precision, closely aligning with the visibility ground truth. This indicates that utilizing our workflow allows for achieving accurate visibility predictions with negligible extra effort, essentially making it a \verb+"+free lunch\verb+"+.

\section{Experiments}

\begin{table*}[ht!]
	\centering
        \resizebox{\textwidth}{!}{
	\begin{tabular}{|c|c|c|c|c|c|c|c|c|c|c|c|}
		\hline
		Method &\multicolumn{3}{|c|}{Car} & \multicolumn{3}{|c|}{Pedestrian} &\multicolumn{3}{|c|}{Cyclist} & mAP & speed\\
		\hline
		\multicolumn{1}{|c|}{ }& Easy & Moderate & hard & Easy & Moderate & hard & Easy & Moderate & hard&   &  \\
		\hline
		PointPillars & 86.79 & 77.12 & 75.59 & 54.52 & 49.19 & 45.33 & 77.32 & 62.01 & 58.10 & 65.10 & 26.79ms  \\
            \hline
		PointPillars + Visibility & 86.79 & 76.72 & 74.19 & 54.28 & 49.82 & 45.34 & 78.63 & 61.22 & 57.32 & 64.92 \textcolor{red}{(-0.18)} & 27.01ms \textcolor{red}{(+0.22ms)} \\
  
		\hline
		\hline
		SECOND & 87.52 & 78.17 & 76.91 & 54.90 & 51.80 & 46.75 & 83.62 & 68.78 & 64.60 & 68.11 & 32.18ms \\
		\hline
		SECOND + Visibility & 88.13 & 78.26 & 76.84 & 57.53 & 53.93 & 48.54 & 81.72 & 66.74 & 61.78 & 68.16 \textcolor{PineGreen}{(+0.05)} & 32.41ms \textcolor{red}{(+0.23ms)} \\
		\hline
	\end{tabular}}
        \caption{Results on KITTI Val 3D detection benchmark with AP. The speed is measured on a Tesla V100 with one GPU 32G.}
        \label{table:tab_1}
\end{table*}

In our experiments, we use the KITTI\cite{geiger2012we} dataset for both training and evaluation, focusing on 3D object detection using only LiDAR sensor data. Our approach leverages two prediction models: the pillar-based PointPillars method~\cite{lang2019pointpillars} and the voxel-based SECOND method~\cite{yan2018second}. 

We aim to assess the impact of incorporating visibility prediction on model perception accuracy, both quantitatively and qualitatively. Additionally, we will compare the time efficiency between traditional and proposed workflows, highlighting the potential benefits of our approach.

\subsection{Model perception accuracy}

In the proposed workflow, the model concurrently predicts a new attribute, i.e. the visibility, alongside other critical object attributes such as classification, size, and orientation. The goal is to integrate visibility prediction with minimal adverse effects on the original model's accuracy and processing speed, effectively making visibility prediction a \verb+"+free lunch\verb+"+.

During our experiments, we evaluated the performance impact of adding visibility prediction to the original model. We compared the original and multi-task learning approaches—where the latter includes visibility prediction—focusing on the accuracy metric mAP and the processing speed (time per frame). The findings are detailed in Table~\ref{table:tab_1}.

Table~\ref{table:tab_1} reveals that incorporating visibility prediction marginally affects the PointPillars model's accuracy negatively, while slightly improving the SECOND model's accuracy. Regarding processing speed, the multi-task learning approach demonstrates a negligible negative impact.

These results suggest that the visibility prediction can be incorporated nearly cost-free, but offering additional benefits for downstream planning tasks and enhancing the redundancy of autonomous driving safety measures.

\subsection{Quantitative and qualitative analysis of predicted visibility}

For the downstream tasks to benefit, it is crucial that the predicted visibility is highly accurate. In our study, since there is no manually labeled ground-truth available for validating the defined visibility, we assess the accuracy of our multi-task predictions by comparing their absolute error against values generated by the proposed algorithm with the traditional workflow. Subsequently, we will present a qualitative analysis of the predicted visibility.

In the quantitative analysis, we compare the predicted visibility derived from multi-task learning against the visibility directly calculated by the algorithm. Given that object detection inherently includes inaccuracies, such as false positives and false negatives, our evaluation of visibility specifically focuses on the bias observed in only true positive predictions.

Here we introduce the visibility accuracy metric, defined as the Absolute Error $AE$, which measures the discrepancy between the predicted visibility $V_{pred}$ and the visibility calculated by the algorithm $V_{algo}$, expressed as follows:

\begin{equation}
    \centering
AE = |V_{pred} - V_{algo}|.
\end{equation}

Table~\ref{table:tab_2} shows the quantitative visibility errors $AE$ on the KITTI validation dataset.

\begin{table}[h]

\begin{center}
\begin{tabular}{|c|c|c|c|}
\hline
Method & \multicolumn{3}{|c|}{Absolute Error (AE)}\\
\hline
& Car & Cyclist & Pedestrian\\

\hline
Random & 0.38 & 0.43 & 0.39 \\
\hline
PointPillars & 0.10 \textcolor{PineGreen}{(-0.28)} & 0.14 \textcolor{PineGreen}{(-0.29)} & 0.30 \textcolor{PineGreen}{(-0.09)} \\
\hline
SECOND & 0.10 \textcolor{PineGreen}{(-0.28)} & 0.14 \textcolor{PineGreen}{(-0.29)} & 0.23 \textcolor{PineGreen}{(-0.16)} \\
\hline
\end{tabular}
\end{center}

\caption{Visibility Absolute Error}

\label{table:tab_2}
\end{table}

\begin{figure*}
      \centering
      \includegraphics[scale=0.105]{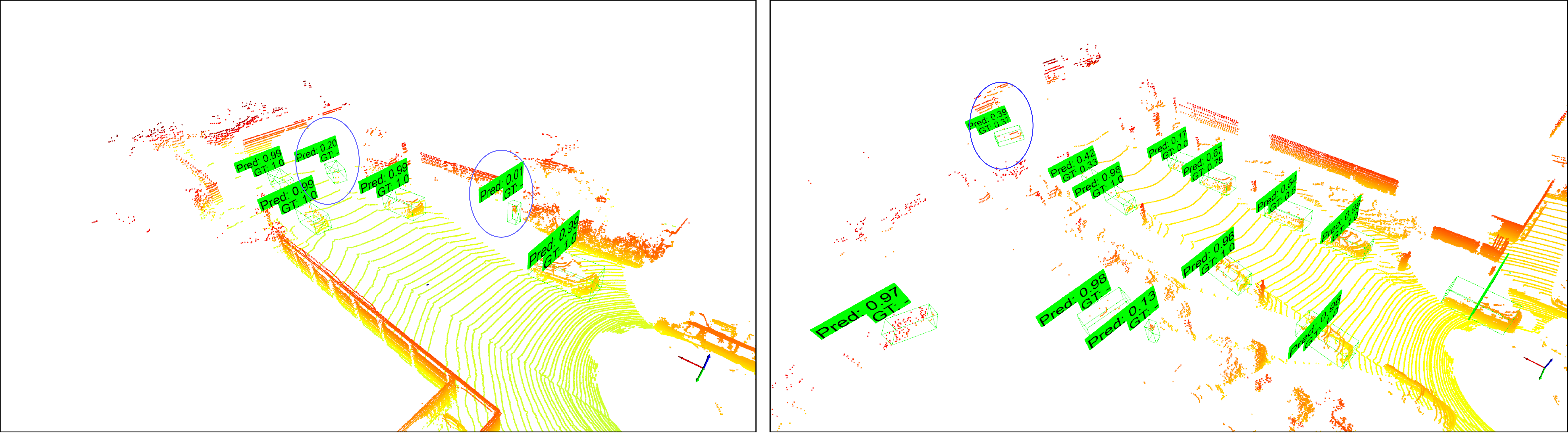}
      \caption{Qualitative examples of predicted Bbox, predicted visibility and ground-truth (GT) visibility, which is derived from algorithm using GT Bbox. If one Bbox is predicted but is not GT, the GT visibility is noted as \textbf{-}. These examples shows that the multi-task learning model has high similar visibility compared to the GT. The three heavily occluded objects within blue circles indicate possibly helpful information for downstream tasks in real-world scenarios.}
      \label{qualitative}
   \end{figure*}

The visibility error range spans from 0 to 1. When employing a random method, visibility predictions are made arbitrarily. Conversely, in the PointPillars and SECOND models, the prediction error markedly decreases, underscoring that these models effectively learn the visibility information through the algorithm.

In the cases of PointPillars and SECOND, the analysis reveals a negative correlation between error magnitude and object size. This means that larger objects, such as cars, tend to have smaller errors compared to smaller objects. This observation is logical, considering that smaller objects offer a limited view—specifically, a reduced solid angle—making them more prone to instability. For instance, the visibility of a pedestrian can drastically drop from 1 to 0.5 as they begin to be occluded, whereas the visibility of a car decreases more gradually under similar conditions.

In qualitative analysis, Figure~\ref{qualitative} presents examples of visibility, demonstrating that the predicted visibility aligns with human perception. This alignment enhances the practical relevance of the findings in the context of driving. Specifically, Figure~\ref{qualitative} highlights three heavily occluded objects within blue circles, illustrating how such \verb+"+free lunch\verb+"+ occlusion information can significantly contribute to the safety of real-time autonomous driving.

\subsection{Extra time consumption in multi-task learning}

A significant contribution of our study is the demonstration that the visibility of additional attributes can be accurately predicted through multi-task learning, effectively turning it into a \verb+"+free lunch\verb+"+ scenario with high accuracy and negligible impact on speed.

In the preceding sections, we have illustrated that the predicted visibility exhibits high precision, particularly with larger objects. This section compares the additional time required for visibility prediction/calculation between the algorithm without multi-task learning, as shown in in Figure~\ref{workflow_traditional}, and our proposed multi-task learning workflow, as shown in Figure~\ref{workflow_new}.

\begin{figure}[thpb]
      \centering
      \includegraphics[scale=0.48]{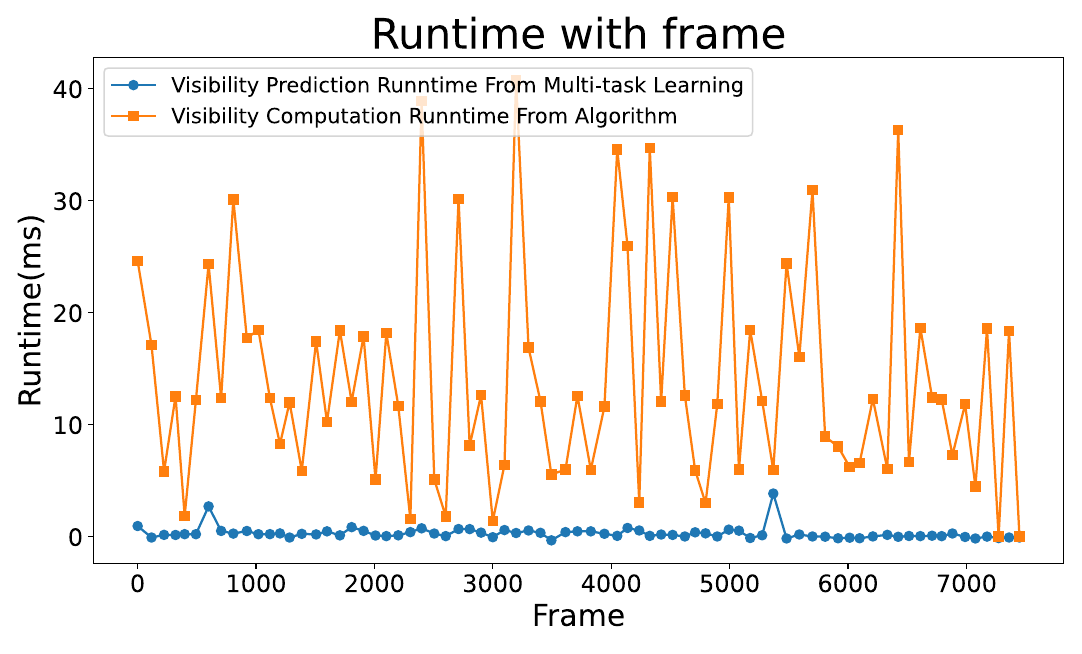}
      \caption{The comparison of visibility prediction runtime between the multi-task learning and algorithm on the KITTI training and validation dataset.}
      \label{visibility-Runntime_with_Bbox }
   \end{figure}

Figure~\ref{visibility-Runntime_with_Bbox } presents a comparison of the additional visibility time between the algorithm and multi-task learning approaches. It is evident that the multi-task learning method (illustrated by the blue plot) significantly reduces the runtime when compared to the algorithm (represented by the orange plot). It is important to note that the runtime mentioned exclude the original model prediction time.

\begin{figure}[thpb]
      \centering
      \includegraphics[scale=0.48]{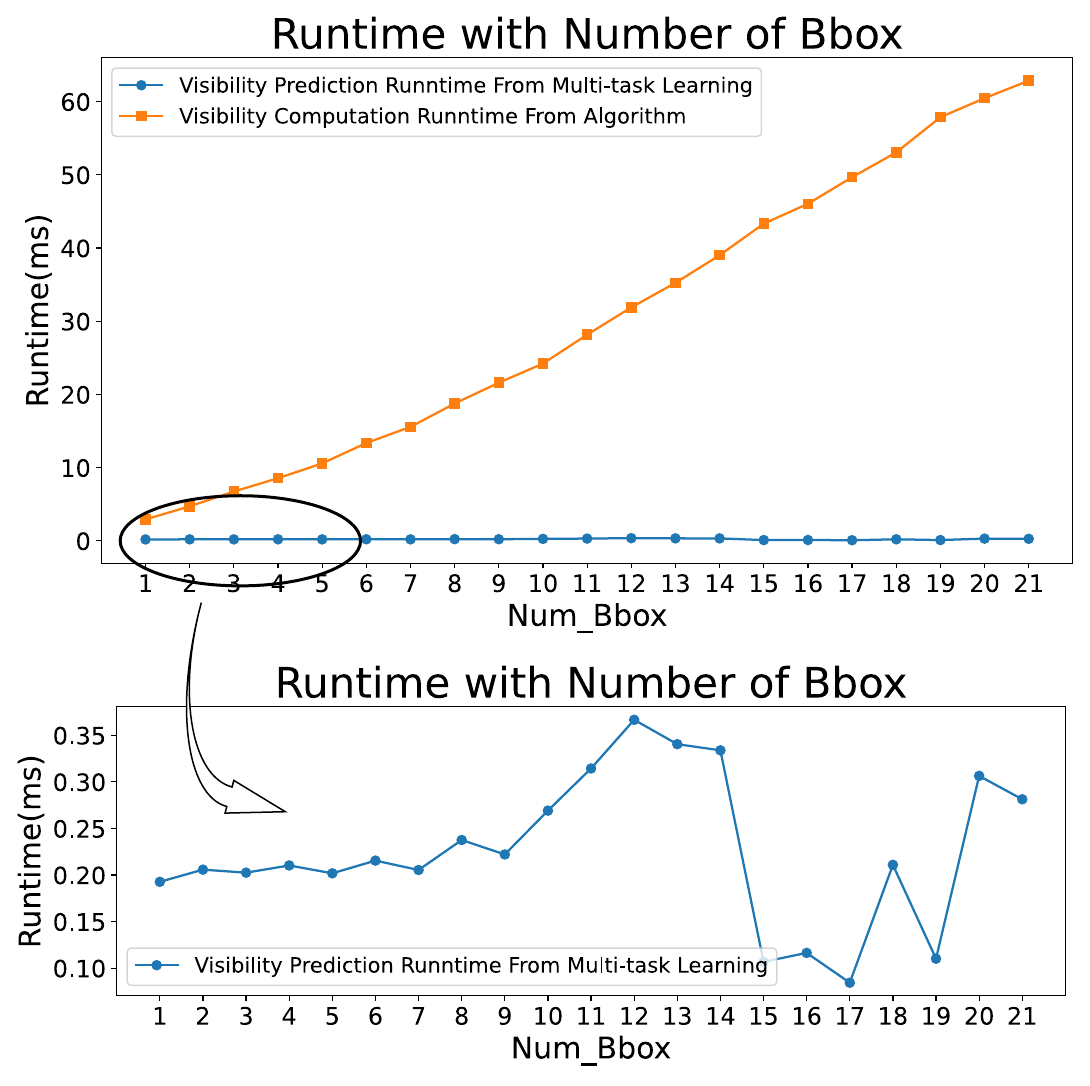}
      \caption{The runtime of visibility predicted by algorithm increases if more Bboxes. However, the prediction time of multi-task learning is robust.}
      \label{visibility-Runntime_with_num}
   \end{figure}

As discussed in subsection~\ref{subsec:algo} and~\ref{subsec:visibility_p}, the algorithm complexity is $O(N^2)$ or $O(NlogN)$ if optimized, while visibility prediction through multi-task learning remains at $O(1)$. We substantiate this claim through experiments. As depicted in Figure~\ref{visibility-Runntime_with_num}, our proposed method demonstrates robustness in speed, even within crowded scenarios.

\section{CONCLUSIONS}

In this paper, we present a novel definition of 3D Bbox visibility alongside an accompanying algorithm. Through the utilization of multi-task learning, we demonstrate that the new attribute of visibility can be acquired with minimal influence on both model accuracy and computational speed.

Given that visibility constitutes just one attribute obtained through our algorithmic approach, our future endeavors will delve into exploring additional possibilities for other attributes attainable via algorithmic means. These tasks, facilitated by multi-task learning, aim to uncover the untapped potential of a singular model, while simultaneously ensuring redundancy of information and bolstering safety standards in autonomous driving scenarios.

\addtolength{\textheight}{-12cm}   




\end{document}